\documentclass[10pt,twocolumn,letterpaper]{article}

\usepackage{iccv}
\usepackage{times}
\usepackage{epsfig}
\usepackage{graphicx}
\usepackage{amsmath}
\usepackage{amssymb}
\usepackage{booktabs}
\usepackage{caption}
\usepackage{floatrow}   
\usepackage{subcaption}
\usepackage{multirow}
\usepackage{pifont}
\usepackage{longtable}

\usepackage{authblk}

\usepackage[pagebackref=true,breaklinks=true,letterpaper=true,colorlinks,bookmarks=false]{hyperref}

\iccvfinalcopy 

\def\iccvPaperID{10259} 

\ificcvfinal\fi

\begin{document}

\title{\texttt{iEdit}: Localised Text-guided Image Editing with Weak Supervision}
\author{Rumeysa Bodur\thanks{Work done during an internship with Amazon.}, 
			Erhan Gundogdu, 
			Binod Bhattarai, \\
			Tae-Kyun Kim,
			Michael Donoser,
			Loris Bazzani\\
			\small{$^1$Imperial College London, UK, $^2$Amazon, $^3$University of Aberdeen, UK, $^4$KAIST, South Korea}		
			
$^2${ {\{eggundog, donoserm, bazzanil\}@amazon.com}}\\
$^1$r.bodur18@imperial.ac.uk, $^3$binod.bhattarai@abdn.ac.uk, $^{1, 4}$kimtaekyun@kaist.ac.kr

}

\maketitle
\ificcvfinal\thispagestyle{empty}\fi

\begin{abstract}
Diffusion models (DMs) can generate realistic images with text guidance using large-scale datasets. However, they demonstrate limited controllability in the output space of the generated images. We propose a novel learning method for text-guided image editing, namely \texttt{iEdit}, that generates images conditioned on a source image and a textual edit prompt. As a fully-annotated dataset with target images does not exist, previous approaches perform subject-specific fine-tuning at test time or adopt contrastive learning without a target image, leading to issues on preserving the fidelity of the source image. We propose to automatically construct a dataset derived from LAION-5B, containing pseudo-target images with their descriptive edit prompts given input image-caption pairs. This dataset gives us the flexibility of introducing a weakly-supervised loss function to generate the pseudo-target image from the latent noise of the source image conditioned on the edit prompt. To encourage localised editing and preserve or modify spatial structures in the image, we propose a loss function that uses segmentation masks to guide the editing during training and optionally at inference. Our model is trained on the constructed dataset with 200K samples and constrained GPU resources. It shows favourable results against its counterparts in terms of image fidelity, CLIP alignment score and qualitatively for editing both generated and real images.

\end{abstract}
\section{Introduction}

\begin{figure}[h!]
\centering
\includegraphics[width=\linewidth]{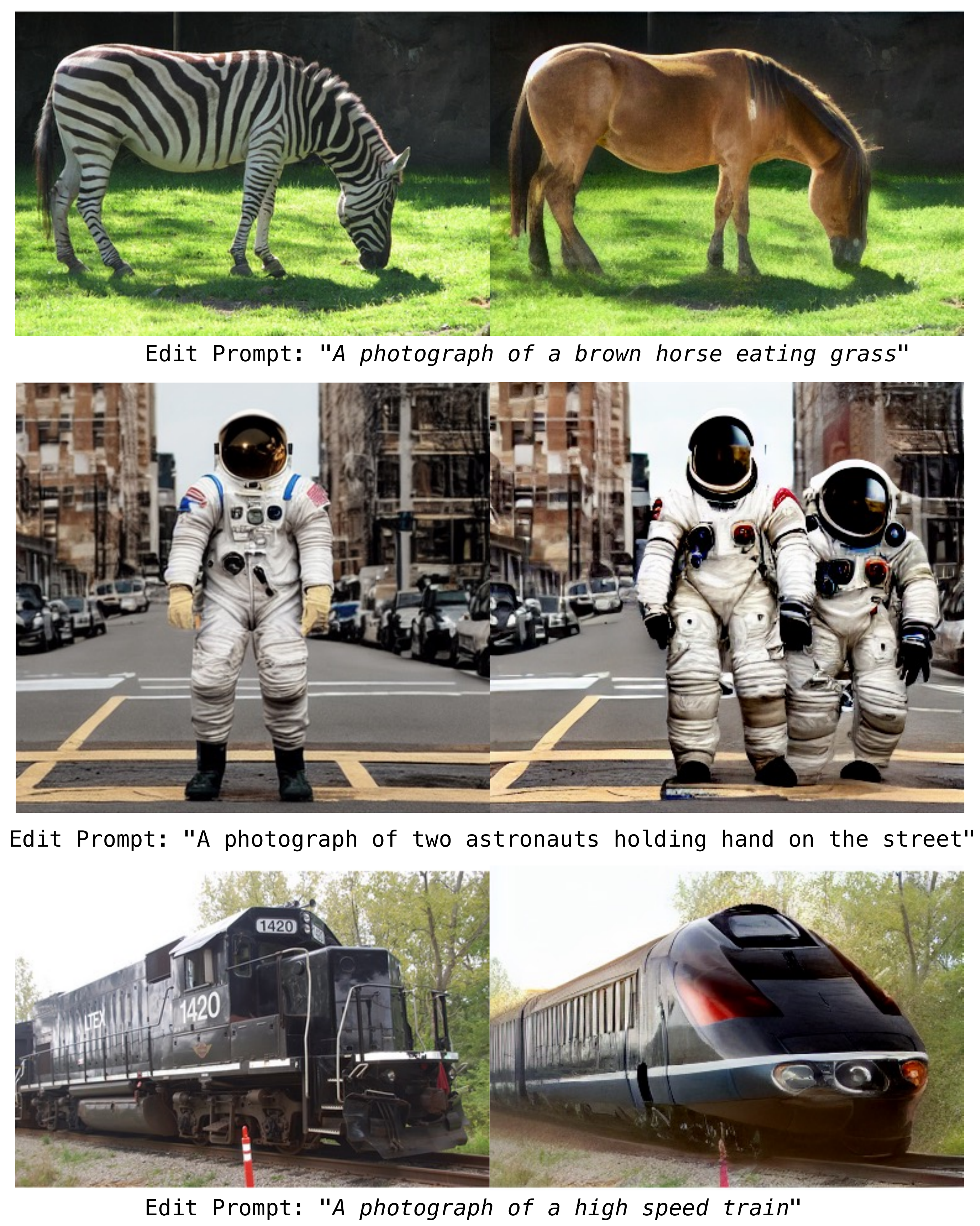}
\RawCaption{\caption{Examples showing that our method can edit images with a textual prompt while preserving image fidelity in the regions not related to the edit.}
\label{figmain:images}}
\end{figure}
Significant progress has been made in developing large-scale text-to-image generative models~\cite{Imagen, ramesh2022dalle, GLIDE, rombach2022sd}, which enable artists and designers to create realistic images without much specialised expertise. 
Existing methods exhibit limited level of controllability, because they are sensitive to the guiding prompt, \emph{i.e.}, a small change in the input text yields a significantly different output image. 
Image editing with generative models increases controllability and help artists and designers make their work personal, creative and authentic.
Recent methods for image editing~\cite{GLIDE, meng2022sdedit, dreambooth, Composer} incorporate various types of inputs (e.g., text, mask, stroke) to reduce diversity in the output space and facilitate the generation of more specific content.

In this paper, we propose \texttt{iEdit}, a framework based on Latent Diffusion Models~\cite{rombach2022sd} (LDMs) for \emph{text-guided image editing}: given a real or generated image and a user-provided textual prompt for editing, we generate a new image which includes targeted and localised modifications as displayed in Fig.~\ref{figmain:images}.
Such editing tasks involve adding or removing objects, changing the appearance of specific regions, or modifying the overall composition of the image while preserving its semantic content.
Text-guided image editing requires the ability to \emph{preserve} the fidelity of the shape, style and semantics for some parts of the image, while \emph{synthesising} realistic modifications that are consistent with the edit prompt for other parts of the image.
To this end, we introduce a method to automatically generate paired datasets specifically suited for image editing, customise LDMs~\cite{rombach2022sd} with the ability of performing editing by aligning the generated images with the target text, and provide shape and location awareness to the method by leveraging segmentation masks.

A supervised approach to train image editing models requires a fully annotated dataset consisting of triples: source image, edit prompt and target image. 
These datasets are expensive to collect and include several challenges for preparing the annotation tasks and guidelines, such as, ambiguity due to vague annotation instructions or potential bias of too specific instructions. 
Hence, we follow the weakly-supervised approach and propose a method to automatically generate editing datasets by semantically pairing images from existing image-caption datasets (e.g., LAION-5B~\cite{schuhmann2022laionb}) using multi-modal embeddings (e.g., CLIP~\cite{CLIP}) to obtain pseudo-target images.
Since captions describe images but not edits, we propose to generate edit prompts by captioning the source image (e.g., BLIP~\cite{li2022blip}) and changing its adjectives and nouns. 
The full process is automatic and enables us to generate a potentially large number of triplets.

\texttt{iEdit} is able to leverage the automatically-constructed dataset and fine-tune LDMs for editing, in contrast with subject-specific or sample-specific training methods~\cite{prompt-to-prompt, meng2022sdedit, BelndedDiffusion, diffedit, diffusionclip}. 
Such methods use a set of user-provided images and target labels/attributes to overfit the models to generate specific content.
However, they lack generalisation capabilities and scalability, i.e., a specific model needs to be fine-tuned for each of the set of target images or attributes.
Moreover, \texttt{iEdit} embeds the semantics of the edit prompt using a contrastive learning loss~\cite{CLIP, diffusionclip} that aligns the generated image and edit prompt. 
To preserve the regions of the source image not mentioned in the edit prompt and apply changes only to the relevant regions, we propose a loss function on masks automatically-extracted from images with an off-the-shelf segmentation model. 

To summarise, the contributions of our paper are the following. 
We present a method to automatically construct paired datasets that are used during training in a weakly-supervised way. 
We present a novel image editing method that is trained to align source and pseudo-target images with edit prompts, and can edit both generated and real images.
We introduce loss functions that use semantic masks to enable localised preservation and synthesis of semantics and regions for editing.
Our method can be trained in a light-weight way by fine-tuning subparts of the network backbone using 2 NVIDIA V100s.
Qualitative and quantitative results demonstrate that our method outperforms state-of-the-art counterparts.

\section{Related Work}
\label{sec:relatedwork}

\paragraph{Text-guided Image Generation.}
DMs \cite{DPM2015, ho2020ddpm} have become the \emph{de-facto} alternative to Generative Adversarial Networks \cite{IanGAN} (GANs) for image generation. 
Stable training allows to increase the capacity of DMs, such as DALL-E 2 \cite{ramesh2022dalle}, GLIDE \cite{GLIDE}, Imagen \cite{Imagen}, and to fairly compare~\cite{dhariwal2021diffusion} with the GAN-based counterparts~\cite{make-a-scene, DM_GAN_CL}. 
LDMs \cite{rombach2022sd} address the computational limitation by working on a latent low-dimensional latent space. 
Similarly, we provide LDMs the ability of image editing, while keeping computational resources constrained (2 NVIDIA V100 GPUs), so that training is more attainable by the research community.

\vspace{-2mm}\paragraph{Text-guided Image Editing with GANs.}
GANs were adopted for image manipulation using text guidance. 
CLIP2StyleGAN \cite{CLIP2StyleGAN}, StyleCLIP \cite{patashnik2021styleclip} and TediGAN \cite{TediGAN} effectively combine StyleGAN \cite{StyleGAN} and CLIP \cite{CLIP} latent embeddings. 
StyleGAN-NADA \cite{StyleGAN-NADA} proposes a text-driven method for out-of-domain generation - one step before free-form text editing. 
ManiGAN \cite{ManiGAN} proposes to train GAN models with multiple stages with text input to edit input images. 
However, GANs are hard to train in a stable way to perform localised editing with large datasets and diversity of input types.
To introduce local edits, Text2Live~\cite{bar2022text2live} proposes to automatically learn an edit layer (similar to masks) that is then combined during image generation.


\vspace{-2mm}\paragraph{Text-guided Image Editing with Diffusion Models.}
DMs can be adapted for text-guided image editing to deal with the challenges of GANs. 
\emph{Prompt-to-prompt} \cite{prompt-to-prompt} proposes cross-attention to align patches to the edit prompt without fine-tuning.
It requires inversion~\cite{dhariwal2021diffusion} to use real images and the resolution of attention maps is lower than masks used in our method.
SDEdit~\cite{meng2022sdedit} uses user guidance, like stroke painting, to generate images and can be conditioned to edit prompts~\cite{diffedit}.
SDEdit is sensitive to the strength parameter of the denoising process that can lead to over-preservation or forgetting the source image.
These conditional DMs modify the image globally, undesirably changing regions that are not mentioned in the edit prompt.

Given that DMs are trained on large datasets that capture diversity and image variability, recent work focuses on fine-tuning on a single image available~\cite{imagic} at test time, a small set of subject-specific images~\cite{dreambooth} or specialises on attribute specific modifications~\cite{dreambooth}.
However, this leads to high computation at test time and limited scalability, since they need to fine-tune a model for each image, set or attribute.
In contrast, our approach requires a single fine-tuning step on the dataset automatically generated for editing.

Other methods like Imagen Editor \cite{EditBench} and Blended Diffusion \cite{BelndedDiffusion} introduce the use of masks for image editing.
However, masks are manually-provided and available at test time.
DiffEdit\footnote{Even if DiffEdit is one of the closest methods to ours, the code/models were not available at the moment of the experiments, so it was not possible to fairly compare.} \cite{diffedit} predicts masks at test time from text-conditioned latent noise differences of the modification in textual descriptions. 
In contrast, our method uses automatically-extracted masks during training and optionally at test time to provide localised properties to the editing framework. 
Recently, Composer \cite{Composer} proposes to decompose images into a set of factors, including masks, and to recompose back to images during training.
Our method is lighter than Composer (1.1B vs 3B parameters) and requires a smaller set of training images (200K vs 60M).

\vspace{-2mm}\paragraph{Training Data for Image Editing.}
Automatic dataset generation for editing was recently explored in InstructPix2Pix \cite{InstructPix2Pix}.
It generates instructions from available image captions using GPT-3 \cite{GPT3} and a pre-trained LDM \cite{rombach2022sd} to generate target images.
In contrast, our dataset construction method is based on retrieval, therefore, lightweight and does not require DMs to build a dataset with synthetic images. 

\section{\texttt{iEdit}}
\label{sec:iedit}
In this section, we first describe our method to construct a dataset with pseudo-target images for text-guided image editing. Next, we provide a brief review of LDMs and then present the proposed image editing method. 
Finally, we extend it to be location aware including segmentation masks and guide the editing process during training and inference.

\subsection{Paired Dataset Construction}
\label{sec:paireddataset}
Most methods discussed in Sec.~\ref{sec:relatedwork} are trained or fine-tuned without a target image, \emph{i.e.}, a ground-truth image after the edit has been applied to the source image.
To train image editing models in a weakly-supervised way, we propose a method to automatically generate pseudo-target images that best match source images and edit prompts.
An ideal image editing dataset should include pairs of images that are nearly identical, except for a change in a specific attribute or object, and both source and target images should have captions that highlight the differences between the pairs.
Due to the lack of such an annotated dataset and since manually-creating one can be an expensive and laborious task, we automatically construct it utilising a publicly-available large-scale image-caption dataset: LAION-5B~\cite{schuhmann2022laionb}. 

\begin{figure}
\begin{center}
\includegraphics[width=\linewidth]{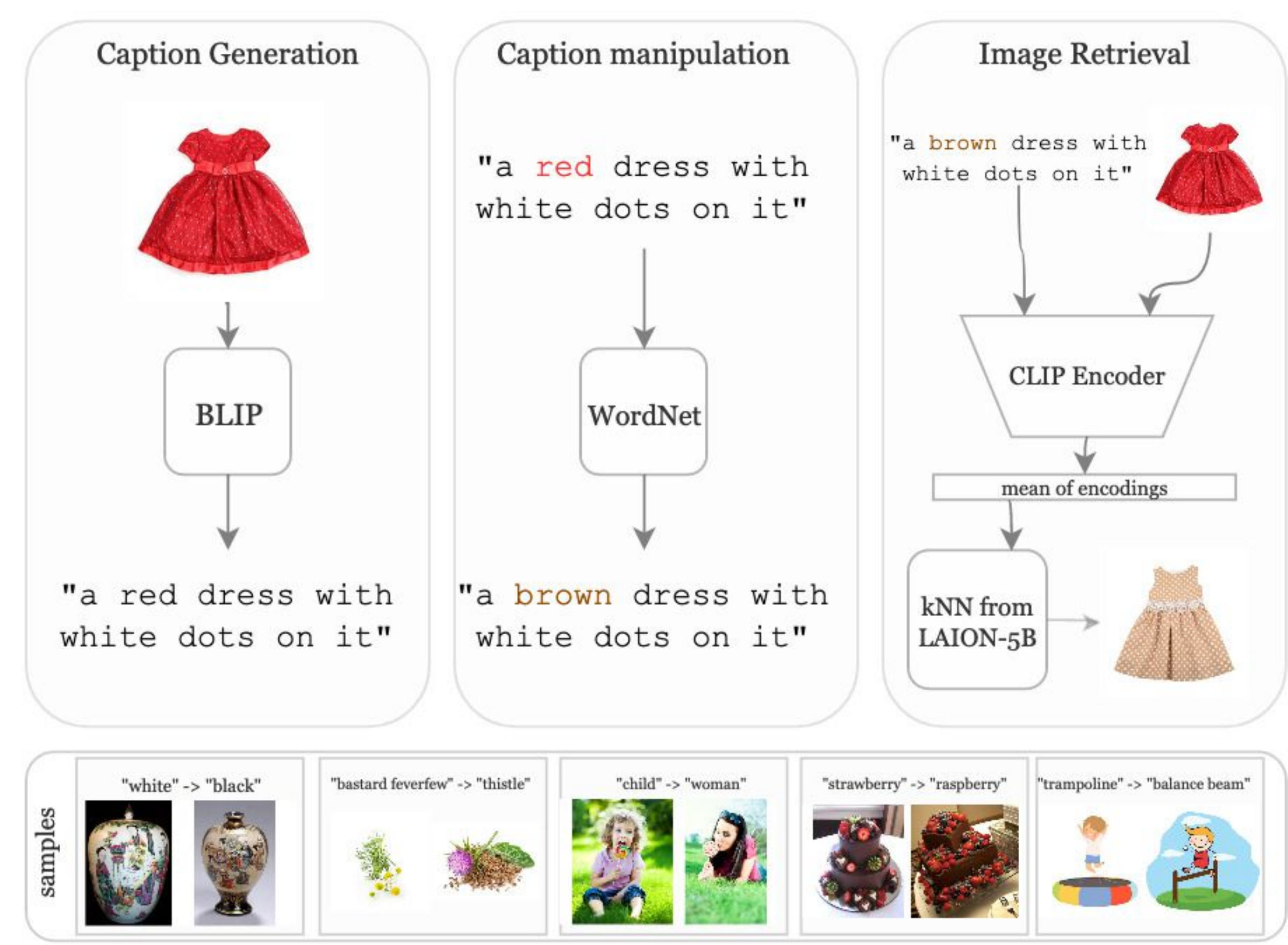}
\end{center}
   \caption{\textbf{Paired dataset construction.} To construct a paired dataset for training, we generate captions with BLIP. Then, captions are manipulated by replacing nouns or adjectives with antonyms or co-hyponyms. We assign the pseudo-target image by using CLIP embeddings of the edit prompt and the source image to retrieve nearest neighbours.}
\label{fig:dataset}
\end{figure}

The source captions in LAION-5B are complex and noisy because they are extracted from the web. 
Therefore, we propose a technique to convert the source captions to a desired edit as depicted in Fig.~\ref{fig:dataset}.
The first step is to generate a simplified version of the caption for each image using the state-of-the-art image captioning method BLIP~\cite{li2022blip}.
Then, the generated captions are manipulated in the second step by replacing an adjective or noun with its antonym or a random co-hyponym using the WordNet library~\cite{fellbaum1998wordnet}, e.g., "a blue and orange \textit{bus} on white background" $\rightarrow$ "a blue and orange \textit{train} on white background". 
The resulting text is our edit prompt.
In the final step, we need to obtain a pseudo-target image, given the source image and the edit prompt.
To this end, we extract CLIP~\cite{CLIP} embeddings of the source image and the edit prompt. 
Finally, we use the mean of these embeddings to retrieve the nearest-neighbour image of each image-prompt pair. 
With this set of operations, we generated a dataset consisting of ${\sim}200K$ image pairs and edit prompts.
Compared with the original captions in LAION-5B, our edit prompts are simpler, shorter and contain only essential information (e.g., the original caption for the red dress in Figure~\ref{fig:dataset} is ''red spring and autumn female short-sleeve princess dress puff skirt one-piece dress costume dress 90 - 135'').
We evaluate our dataset in Sec.~\ref{sec:experiments} and provide examples in Figure~\ref{fig:dataset}. 

\subsection{Background: Latent Diffusion Model}
DMs are a class of probabilistic generative models that consist of a forward diffusion process and a reverse process. In the forward diffusion process of Denoising Diffusion Probabilistic Models (DDPM)~\cite{ho2020ddpm}, the training data is corrupted by adding Gaussian noise and a parameterised model, \emph{e.g.}, a neural network, is used to reverse this process.







In LDMs~\cite{rombach2022sd}, an auto-encoder is trained to transform images into a lower-dimensional latent space to reduce the computational complexity and increase the inference speed of DMs. 
An encoder $\mathcal{E}$ compresses the input image $x$ to a smaller 2D latent vector $z=\mathcal{E}(x)$. 
Both forward diffusion and reverse denoising process happen in the latent space, and then a decoder $\mathcal{D}$ reconstructs the image from the latent vector, $x~=\mathcal{D}(z)$. 

The denoising step can be \emph{conditioned} on text, an image or other data $y$. 
A domain-specific encoder $\tau_{\theta}$ projects the conditioning input to an intermediate representation for each type of conditioning. An encoding of the conditioning data is exposed to the denoising U-Nets~\cite{ronneberger2015unet} via a cross-attention mechanism or concatenation to $z_T$. The loss for conditional LDM is:
\begin{equation}
{\mathcal{L}_{LDM}} = \mathbb{E}_{\mathcal{E}(x),y, \epsilon\sim\mathcal{N}(0,1),t}[||\epsilon - \epsilon_{\theta}(z_{t},t,\tau_{\theta}(y))||^2,
\label{eqn:ldm_simple_loss}
\end{equation}
where $\epsilon_{\theta}$ is a U-Net conditioned to the step $t$ uniformly sampled from $\{1,2,...,T\}$. The network parameters $\theta$ are optimised to predict the noise ${\epsilon_1}{\sim}\mathcal{N}(0,1)$ that is used for corrupting the encoded version of the input image. At inference time, the trained model is sampled by iteratively denoising ${z}{\sim}\mathcal{N}(0,1)$ using the deterministic DDIM~\cite{ddim}.

\subsection{\texttt{iEdit} with Weak Supervision}
We present here our method to fine-tune the LDM \cite{rombach2022sd} for image editing using the weakly-supervised dataset described in Sec.~\ref{sec:paireddataset}.
Let $x_1$ and $x_2$ be the source and pseudo-target images, respectively, and $y_2$ is the edit prompt of $x_2$, derived from $y_1$ with caption $x_1$ as detailed in Sec.~\ref{sec:paireddataset}. We first obtain the noisy image $z_t$ by adding noise $\epsilon_1$ to $z_1 :=\mathcal{E}(x_1)$, the source image encoded in the latent space: 
\begin{equation}
{z_t} = \sqrt{\alpha_{t}}z_1 + \sqrt{1-\alpha_{t}}\epsilon_1,
\label{eqn:ief_noising}
\end{equation}
where ${\epsilon_1}{\sim}\mathcal{N}(0,1)$, and $\alpha_{t}$ is the Gaussian transition sequence following the notation in \cite{song2021ddim}.
We consider $z_t$ as the noisy version of both the source and target image, since our aim is to generate the target image from $z_t$. We, then, calculate the ground truth noise for reconstructing $z_2 := \mathcal{E}(x_2)$, which is the target image encoded in the latent space:
\begin{equation}
\epsilon_2 =  \frac{{z_t -\sqrt{\alpha_{t}}z_2}}{\sqrt{1-\alpha_{t}}}.
\label{eqn:ief_gt_noise}
\end{equation}
Hereby, we modify the objective in Eq.~\ref{eqn:ldm_simple_loss} to minimise the L2 loss between this ground truth noise, $\epsilon_2$, and the noise predicted by the network, $\epsilon_\theta$, given the edit prompt:
\begin{equation}
\mathcal{L}_{paired} = \mathbb{E}_{\mathcal{E}(x),y_{2}, \epsilon_2,t}[||\epsilon_2 - \epsilon_{\theta}(z_{t},t,\tau_{\theta}(y_{2}))||^2]
\label{eqn:ief_simple}
\end{equation}
To further encourage the generated image to be aligned with the edit prompt, we introduce a global CLIP loss~\cite{patashnik2021styleclip} between the edit prompt and the generated image $\hat{x}_{1}$:
\begin{equation}
\mathcal{L}_{global}(\hat{x}_{1}, y_{2}) = D_{CLIP}(\hat{x}_{1}, y_{2}) 
\label{eqn:loss_global_clip}
\end{equation}
where $\hat{x_{1}}$ is obtained with the decoder $\mathcal{D}$ as follows:
\begin{equation}
\hat{x_{1}}=\mathcal{D}(\hat{z}_1)
\label{eqn:ief_rec1}
\end{equation}
\begin{equation}
\hat{z}_1 =  \frac{{z_t -\sqrt{1-{\alpha}_{t}}\epsilon_{\theta}(z_{t},t,\tau_{\theta}(y_{2}))}}{\sqrt{{\alpha}_{t}}}.
\label{eqn:ief_rec2}
\end{equation}
 
With higher noise levels the reconstruction is more successful and therefore $L_{paired}$ gives more reliable results. On the other hand, CLIP loss gives more reliable results on low noise levels as CLIP embeddings are ideal for noise-free inputs. Thus, we set inversely proportional weights for these losses based on the noise level $t$ as $(1-\frac{t}{T}) \mathcal{L}_{global} + \frac{t}{T} \mathcal{L}_{paired}$, where $T$ is the maximum number of noise steps.

\subsection{\texttt{iEdit} with Location Awareness}

\begin{figure}
\begin{center}
\includegraphics[width=\linewidth]{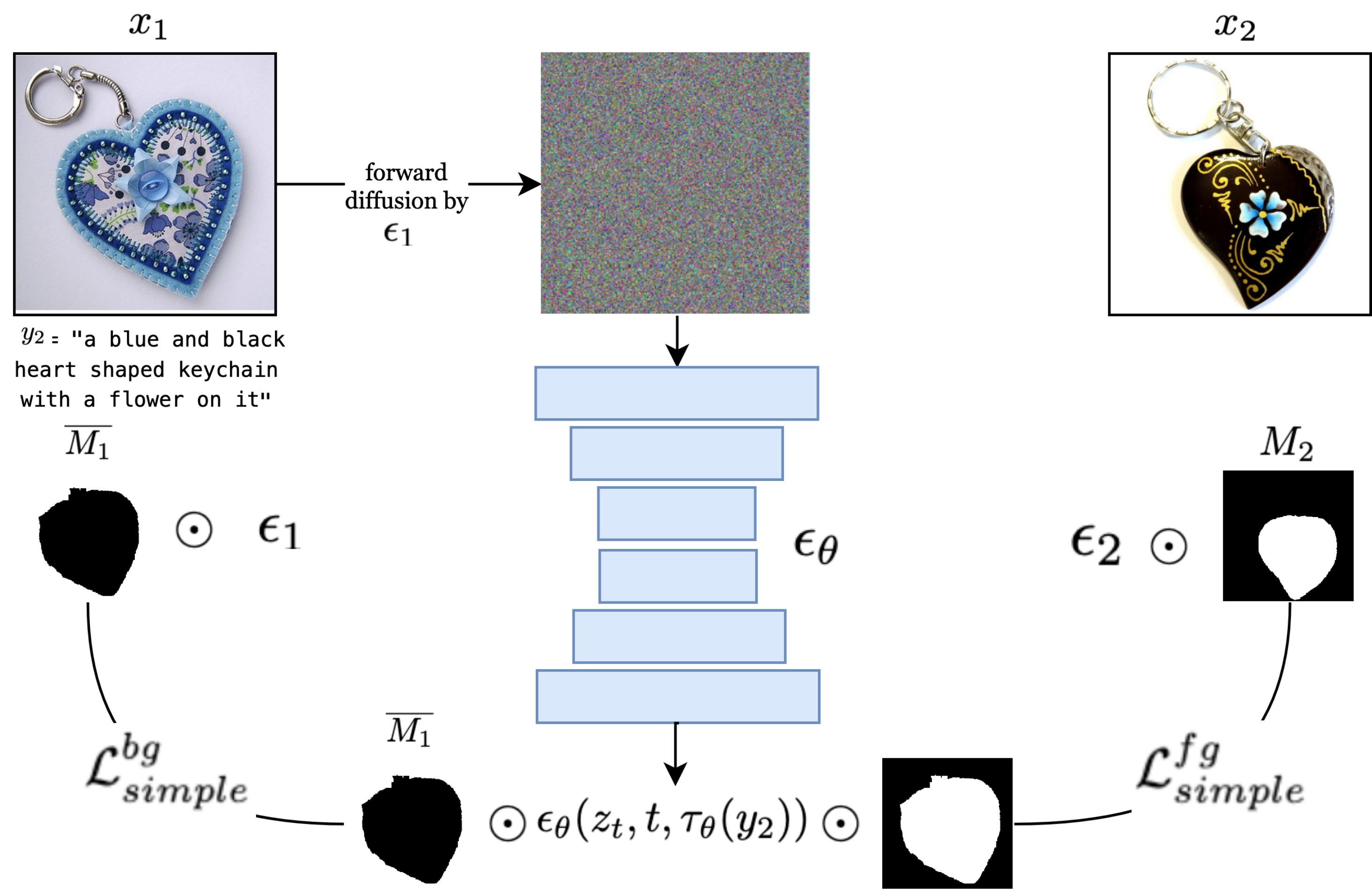}
\end{center}
   \caption{\textbf{Proposed image editing framework.} Our framework takes as input the image pairs from the constructed dataset, their corresponding masks and the edit prompt. We optimise $\epsilon_\theta$ to predict the noises $\epsilon_1$ and $\epsilon_2$ for background and foreground, respectively. $\odot$ denotes element-wise multiplication of the image and mask.}
\label{fig:short}
\end{figure}

We extend \texttt{iEdit} to incorporate masks from images in order to enable localised image editing and better align source and pseudo-target images. 
Specifically, we introduce the use of masks during training to guide the learning process and optionally during inference to generate localised edits. To obtain the masks, we use CLIPSeg~\cite{lueddecke2022clipseg}, a state-of-the-art method that generates image segmentations conditioned to text prompts at test time. We take the differences between the BLIP-generated source caption and the edit prompt along with the noun it describes in case it is an adjective. 
This difference prompt $y_1^{diff}$ (or $y_2^{diff}$) and the corresponding image $x_1$ (or $x_2$) are fed into the CLIPSeg model to obtain masks for both source and target images, $M_1$ and $M_2$, respectively.

\vspace{-2mm}\paragraph{Training with Masks.}
{In Figure~\ref{fig:short}}, we visualise an overview of our method for incorporating masks during training. To enable localised edits, we modify the optimisation process to predict the target noise, $\epsilon_2$, on the masked region and the source noise, $\epsilon_1$, on the inverse mask region, rather than optimising the network to predict only the target noise. Specifically, Equation \ref{eqn:ief_simple} becomes:
\begin{equation}
\mathcal{L}_{mask} = \mathbb{E}_{\mathcal{E}(x),y_{2}, {\epsilon_2},t} \mathcal{L}_{mask}^{fg} + \mathbb{E}_{\mathcal{E}(x),y_{2}, {\epsilon_1},t} \mathcal{L}_{mask}^{bg}
\label{eqn:ldm_simple_loss_mask_main}
\end{equation}
where $\mathcal{L}_{mask}^{fg}$ and $\mathcal{L}_{mask}^{bg}$ represent the foreground and background loss terms as follows:
\begin{equation}
\begin{split}
\mathcal{L}_{mask}^{fg} = [||\epsilon_2 \odot M_2 - \epsilon_{\theta}(z_{t},t,\tau_{\theta}(y_{2}))\odot M_2||^2] \\
\mathcal{L}_{mask}^{bg} = [||\epsilon_1 \odot \overline{M_1} - \epsilon_{\theta}(z_{t},t,\tau_{\theta}(y_{2}))\odot \overline{M_1}||^2]
\end{split}
\label{eqn:ldm_simple_loss_mask}
\end{equation}
where $\overline M_i$ denotes the inverse of the mask, and $\odot$ represents element-wise multiplication.

In addition to $\mathcal{L}_{mask}$, we also employ two additional losses, namely a perceptual loss~\cite{johnson2016perceptual} and a localised CLIP loss, to better translate according to the edit prompt. The perceptual loss is used to ensure that the edited image has a similar visual appearance to the target image in their masked regions. Specifically, we use a pre-trained VGG network~\cite{simonyan15vgg} $V(\cdot)$ to extract features from the edited image $\hat{x}_1$ and the target image $x_2$ and minimise the mean squared error between their feature representations:
\begin{equation}
\mathcal{L}_{perc} = \mathbb{E}_{\hat{x}_1,x_2} [||V(\hat{x}_1 \odot M_1) - V(x_2 \odot M_2)||^2].
\label{eqn:perceptual_loss}
\end{equation}
The localised CLIP loss aims to ensure that the masked area of the generated image is coherent with the difference of the edit prompt with the source caption, $y^{diff}_2$, in terms of semantic content. This loss is defined as:
\begin{equation}
\mathcal{L}_{loc}(\hat{x}_1, y^{diff}_2) = D_{CLIP}(\hat{x}_1 \odot M_1, y^{diff}_2).
\label{eqn:loss_localised_clip}
\end{equation}


In summary, our final loss for fine-tuning \texttt{iEdit} is:
\begin{equation}
(1-\frac{t}{T}) (\mathcal{L}_{loc} + \mathcal{L}_{global}) + \frac{t}{T} \mathcal{L}_{mask} + \lambda_{perc} \mathcal{L}_{perc}.
\label{eqn:final_loss}
\end{equation}
\paragraph{Inference with Masks.}
In SDEdit~\cite{meng2022sdedit}, during inference, random noise is added to the input image, and this corrupted image is denoised with the trained model to generate the edited version. We extend this approach by incorporating masks in order to generate localised edits while preserving the visual content of the source image in the inverse mask region. Given an input image, an edit prompt, and which term in the edit prompt describes the edit area, we apply Gaussian noise corruption to the input image for a certain number of iterations defined by a sampling ratio. We vary this hyper-parameter between $0.6-0.8$ throughout our experiments for both SDEdit and our method, where $0.0$ and $1.0$ correspond to using the input image or a random Gaussian noise as input. 
At each DDIM~\cite{song2021ddim} sampling step, we replace the pixel values of the reconstructed image (\emph{i.e.}, $\tilde{z}_t$) in the inverse mask region $\overline{M}$ with their corrupted version of the input image, $z_t$, at the given time step $t$, as
\begin{equation}
\tilde{z}_t = \overline{M} \odot z_t + M \odot \tilde{z}_t.
\label{eqn:mask_inf}
\end{equation}
The inference with this additive masking described above helps to preserve the details of original latent input $z_1$, progressively at each noise level $t$. Finally, we follow the decoding step as in Eq.~\ref{eqn:ief_rec1} to obtain the final image $\hat{x_{1}}$.


\section{Experiments}
\label{sec:experiments}

\begin{figure*}
\begin{center}
\includegraphics[width=0.95\linewidth]{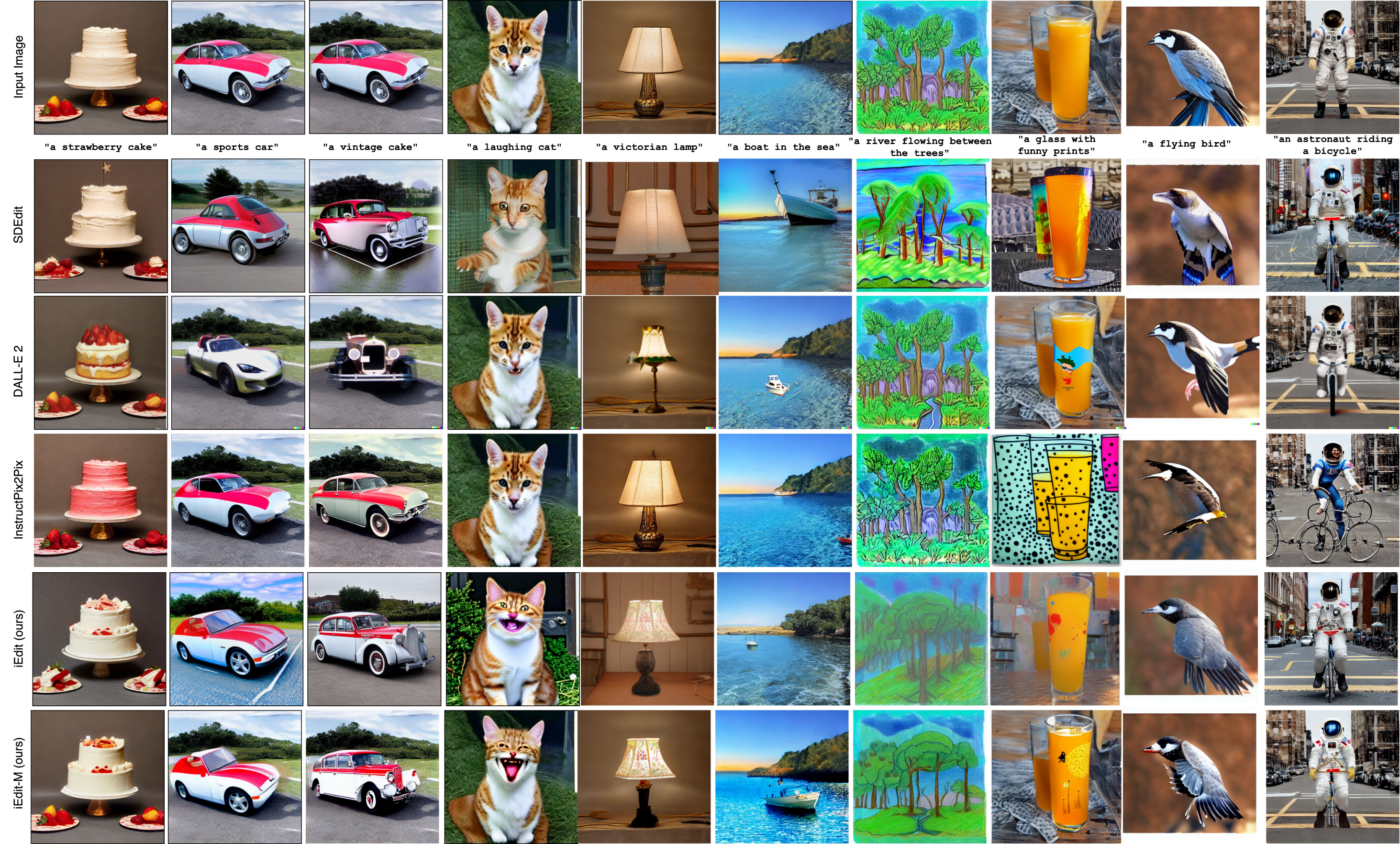}
\end{center}
   \caption{\textbf{Comparison to state-of-the-art on generated images.} Our method produces results with higher fidelity to the source image and the edit prompt compared to SDEdit~\cite{meng2022sdedit}, DALL-E 2~\cite{ramesh2022dalle} and InstructPix2Pix~\cite{InstructPix2Pix}.}
\label{fig:qual_gen}
\end{figure*}

In this section, we perform a thorough comparison of \texttt{iEdit} with state-of-the-art methods by quantitative and qualitative analysis on generated and real source images. 

\subsection{Experimental Setup}
\label{ssec:experimentalsetup}

\paragraph{Training Setting.} We use the LDMs~\cite{rombach2022sd} pre-trained on LAION-5B~\cite{schuhmann2022laionb} with the Stable Diffusion checkpoint v1.4\footnote{\url{https://github.com/CompVis/stable-diffusion}}.
We fine-tune \texttt{iEdit} for ${\sim}{10,000}$ steps on 2 16GB NVIDIA Tesla V100 GPUs for $384 \times 384$ resolution with a batch size of 1 and a learning rate of $2\times10^{-4}$. 
To effectively fine-tune on such a constrained computational setup, we update the input and middle layers of the UNet in an alternate way. 
Following~\cite{tumanyan2022plugandplay, diffedit}, we set the classifier-free guidance scale to $7.5$.
Inference for $4$ possible editing results per image takes around ${\sim}10$ seconds. 

\vspace{-2mm}\paragraph{Datasets.} We fine-tune \texttt{iEdit} with the paired dataset constructed from LAION-5B~\cite{schuhmann2022laionb} as described in Sec.~\ref{sec:paireddataset}. 
For evaluation with generated images, we use LDM~\cite{rombach2022sd} to construct $70$ image-prompt pairs, with $20$ distinct generated images. 
For evaluation with real images, we build $52$ image-prompt pairs consisting of $30$ distinct images from COCO~\cite{lin2014coco}, ImageNet~\cite{deng2009imagenet} and AFHQ~\cite{choi2020stargan}.

\vspace{-2mm}\paragraph{Comparison to Other Methods.} 
We compare to the SDEdit~\cite{meng2022sdedit} extension of Stable Diffusion since we use a similar diffusion process and to recent state-of-the-art methods, such as DALL-E 2\footnote{Web UI accessed in March 2023: \url{https://openai.com/product/dall-e-2}.}~\cite{ramesh2022dalle} and InstructPix2Pix\footnote{Web UI accessed in March 2023: \url{https://huggingface.co/spaces/timbrooks/instruct-pix2pix}.}~\cite{InstructPix2Pix}. 
In order to adapt DALL-E 2 for image editing, we manually provide ground-truth masks for each test image during evaluation. 

\vspace{-2mm}\paragraph{Evaluation Metrics.} 
In text-based image editing, we have to evaluate with respect to two dimensions: 1) how well the method synthesises the regions of the image explicitly mentioned in the edit prompt and 2) the quality of preservation of the rest of the image not mentioned in the prompt.
We adapted the Structural Similarity Index (SSIM)~\cite{wang2004ssim} to capture this intuition as: the SSIM score on the edited area (SSIM-$M$) and on the rest of the image (SSIM-$\overline{M}$) by using manually created ground-truth masks.
In addition, we use CLIPScore~\cite{hessel2021clipscore} to measure the alignment between the edit prompt and the edited image by computing the cosine similarity between their embeddings generated by the CLIP model. 
CLIPScore has been shown in~\cite{EditBench} to be reliable due to its high agreement with human judgement.
We compute the Fréchet Inception Distance (FID)~\cite{seitzer2020fid} to evaluate the quality and fidelity of generated images.

\subsection{Qualitative Evaluation}

\begin{figure*}
\begin{center}
\includegraphics[width=0.95\linewidth]{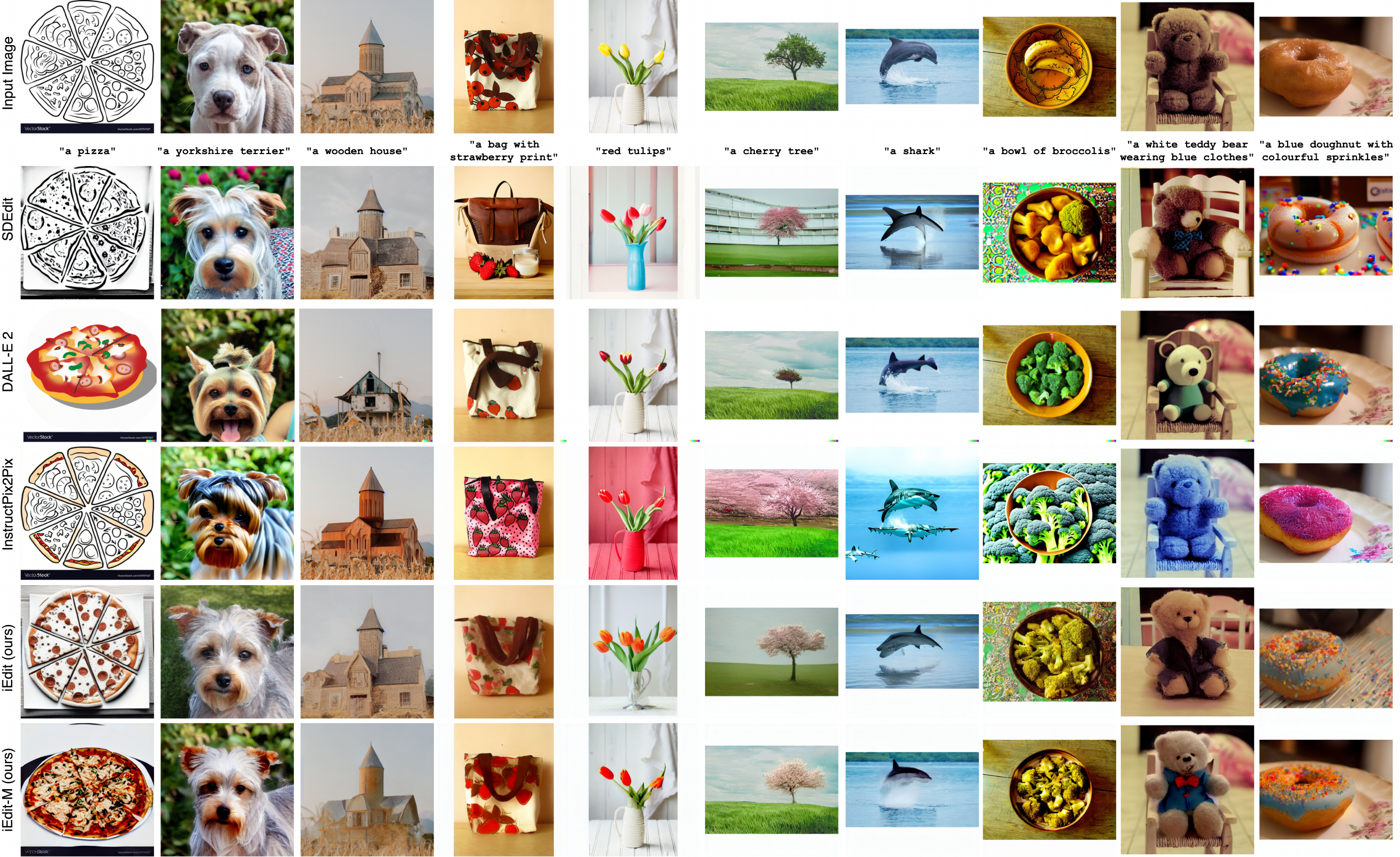}
\end{center}
   \caption{\textbf{Qualitative Results on Real Images.} \texttt{iEdit} outperforms compared methods showing high fidelity to the edit prompt and the input image.}
\label{fig:qual_real}
\end{figure*}

\paragraph{Editing Generated Images.} 
Fig.~\ref{fig:qual_gen} compares  \texttt{iEdit} (our method) and \texttt{iEdit}-$M$ (with predicted masks at inference) with SDEdit~\cite{meng2022sdedit}, DALL-E 2~\cite{ramesh2022dalle}, InstructPix2Pix~\cite{InstructPix2Pix} on the generated images of the evaluation set as explained in Sec.~\ref{ssec:experimentalsetup}. 
We observe that SDEdit~\cite{meng2022sdedit} often fails to perform the desired edit, e.g., \textit{"laughing cat"} in col.~4, \textit{"river"} in col.~7.
It also does not preserve fidelity, e.g., background in \textit{"ship in the sea"} in col.~6 and attributes of the bird in col.~9. 
This is due to SDEdit's trade-off between preservation of the input and editing according to the edit prompt. 
DALL-E 2~\cite{ramesh2022dalle} shows better fidelity to the edit prompt, but it ignores the prior information regarding the object to edit from the source image.
It produces results that completely replace the edited object with a visually different one, e.g., \textit{"sports car"} in col.~2 and \textit{"strawberry cake"} in col.~1. 
The user-provided masks in DALL-E 2 help localising the edit, but its in-painting nature ignores the structure of the edited region from the source images. 
Furthermore, this causes the cases that do not blend well with the general structure of the image, e.g. the tale and wings of the bird in col.~9 and the legs of the astronaut in col.~10 look unrealistic. 
InstructPix2Pix displays high fidelity to both source image and edit prompt, however, it affects the whole layout in some cases, e.g., \textit{"a vintage car"} in col.~3 produces a vintage-style image (including background), and the image in col.~8 results in dots.
This is a limitation~\cite{tumanyan2022plugandplay} of InstructPix2Pix inherited from Prompt-to-Prompt~\cite{prompt-to-prompt}, used to generate the training set. 
Compared to these methods, \texttt{iEdit} performs more successful localised edits, while preserving the background, with small undesired changes. 
This is alleviated with the use of predicted masks in \texttt{iEdit}-$M$ at inference.

\vspace{-2mm}\paragraph{Editing Real Images.} 
One of the main advantages of our method is the ability to edit real images without inversions.
Fig.~\ref{fig:qual_real} shows a qualitative comparison on the test set of real images mentioned in Sec.~\ref{sec:experiments}.
In accordance with the observations on editing generated images, SDEdit~\cite{meng2022sdedit} achieves low fidelity to edit prompt or source image, e.g., edits in \textit{"pizza"} in col.~1 and \textit{"strawberry prints"} on col.~4, or high deviation from the source image, e.g., \textit{"red tulips"} in col.~5 and \textit{"a cherry tree"} in col.~6. 
DALL-E 2~\cite{ramesh2022dalle} yields results that are inconsistent with the source or look unnatural, e.g. \textit{"wooden house"} in col.~3 and \textit{"white teddy bear wearing blue clothes"} in col.~9. 
The limitation of InstructPix2Pix~\cite{InstructPix2Pix} produces results not complying with the style of the source image (\textit{"bag with strawberry prints"} in col.~4) or affecting the whole image (\textit{"cherry tree"} in col.~6). 
It also fails when multiple changes are requested at once, e.g. \textit{"white teddy bear wearing blue"} in col.~9. 
\texttt{iEdit} shows high fidelity to the source image and the edit prompt on real images as well, where undesired background changes are averted with the introduction of masks in \texttt{iEdit}-$M$.
\begin{table*}[t]
    \begin{subtable}[h]{0.45\textwidth}
        \centering
        \small
        \resizebox{\linewidth}{!}{
        \begin{tabular}{ccccc}
         \toprule[0.5pt] 
         \textbf{Method} & \textbf{CLIPScore (\%) $\uparrow$} & \textbf{FID$\downarrow$} & \textbf{SSIM-$M (\%) $} & \textbf{SSIM-$\overline{M}(\%) \uparrow$}\\
         \midrule[0.5pt]
    
         SDEdit~\cite{meng2022sdedit} & 62.58 & 171 & 82.44 & 50.64 \\ 
         DALL-E 2~\cite{ramesh2022dalle} & 65.44 & 143 & {82.45} & \textbf{94.76}  \\
         InstructPix2Pix~\cite{InstructPix2Pix} & 65.12 & \textbf{108} & {88.62}  & 76.43 \\
         \hline
         \texttt{iEdit} (ours) & \underline{65.76} & 158 & 82.70 & 52.02 \\
         \texttt{iEdit}-$M$ (ours) & \textbf{66.36} & \underline{114} & 83.08 & \underline{78.18} \\
         
         \bottomrule[0.5pt] 
    \end{tabular}
    }
    \caption{Results on generated images using LDM~\cite{rombach2022sd}.}

    \label{tab:quantitative_results_fake}
    \end{subtable}
    \hfill
    \begin{subtable}[h]{0.45\textwidth}
    \centering
    \small
    \resizebox{\linewidth}{!}{
    \begin{tabular}{ccccc}
         \toprule[0.5pt] 
        \textbf{Method} & \textbf{CLIPScore (\%) $\uparrow$} & \textbf{FID$\downarrow$} & \textbf{SSIM-$M (\%)$} & \textbf{SSIM-$\overline{M}(\%) \uparrow$}\\
         \midrule[0.5pt]
    
         SDEdit~\cite{meng2022sdedit} &  65.84 & 180 & 74.36 & 64.60 \\ 
         DALL-E 2~\cite{ramesh2022dalle} & 65.46 & 162 & {74.41} & \textbf{93.97}  \\
         InstructPix2Pix~\cite{InstructPix2Pix} & 66.91 & \textbf{145} & {80.59} & 79.92 \\
         \hline
         \texttt{iEdit} (ours) & \underline{67.02} & 166 & 74.59 & 70.09  \\
         \texttt{iEdit}-$M$ (ours) & \textbf{67.44}  & \underline{147} & 74.98 & \underline{80.44}  \\
         
         \bottomrule[0.5pt] 
    \end{tabular}
    }
    \caption{Results on real images from COCO~\cite{lin2014coco}, ImageNet~\cite{deng2009imagenet} and AnimalFaces-HQ~\cite{choi2020stargan}. }
    \label{tab:quantitative_results_real}
     \end{subtable}
     \caption{
     Quantitative results on (a) generated and (b) real images. Best numbers are marked in \textbf{bold}, second best \underline{underlined}.}
     \label{tab:temps}
\end{table*}
\begin{table*}[t]
    \centering
    \footnotesize
    \resizebox{0.7\linewidth}{!}{
    \begin{tabular}{cc|cccc}
     \toprule[0.5pt] 
     \multicolumn{2}{c}{\textbf{Ablation Settings}} & \multicolumn{4}{c}{\textbf{Scores}} \\
     \textbf{Losses} & \textbf{Fine-tuning Dataset} &   \textbf{CLIPScore (\%) $\uparrow$} & \textbf{FID$\downarrow$} & \textbf{SSIM-$M (\%)$} & \textbf{SSIM-$\overline{M}(\%) \uparrow$}\\
     \midrule[0.5pt]

      $\mathcal{L}_{global} + \mathcal{L}_{pairs}$  & LAION-caption-200K &  65.95 & 158 &  78.84 & 55.14 \\
     $\mathcal{L}_{global} + \mathcal{L}_{LDM}$+$\mathcal{L}_{loc}$ &LAION-edit-200K    & 65.62 & 156 & 79.61 & 62.08 \\
     $\mathcal{L}_{global} + \mathcal{L}_{LDM}$+$\mathcal{L}_{loc}$+$\mathcal{L}_{perc}$ &LAION-edit-200K   & 65.64 & 153 &  79.75 & \underline{62.20}  \\
      $\mathcal{L}_{global} + \mathcal{L}_{mask}$+$\mathcal{L}_{loc}$+$\mathcal{L}_{perc}$ &LAION-edit-200K  &  \underline{66.09} & \underline{146} & 79.31 & {59.54} \\
      $\mathcal{L}_{global} + \mathcal{L}_{mask}$+$\mathcal{L}_{loc}$+$\mathcal{L}_{perc}$ +Masked Inference &LAION-edit-200K  & \textbf{66.97} &  \textbf{128} & 79.65 & \textbf{77.99} \\
     
     \bottomrule[0.5pt] 
    \end{tabular}
    }
    \caption{Ablation study of \texttt{iEdit} on our loss functions and datasets.}
    \label{tab:quantitative_ablation}
\end{table*}

\subsection{Quantitative Evaluation}
Table~\ref{tab:quantitative_results_fake} shows a comparison of state-of-the-art methods and \texttt{iEdit} for editing generated images in our evaluation set. 
In terms of CLIPScore, \texttt{iEdit}-$M$ outperforms all compared methods, which means high alignment and fidelity between the generated image and the edit prompt.
\texttt{iEdit}-$M$ is better by $+3.78\%$ and $+1.24\%$ compared to SDEdit and InstructPix2Pix, respectively.  
Note that CLIPScore is one of the main metrics for editing, since it was shown to have high agreement with human judgement~\cite{EditBench}.
We further measure FID, where our method is the second best and is marginally outperformed by InstructPix2Pix.

As SSIM-$M$ measures the SSIM score of the edited area, its high values do not imply a better result, because the edited area could be too similar to that of the source image.
In editing, we want the SSIM-$M$ to be a trade-off between changing the source image accordingly to the edit prompt and not changing it drastically.
So, we attribute the high SSIM-$M$ of InstructPix2Pix to the fact that generated images are too similar to source images and might indicate that the editing was not fully successful. 
The lower values for \texttt{iEdit}(-$M$) implies more changes in the foreground mask, that combined with the higher CLIPScore and the qualitative results in Fig.~\ref{fig:qual_gen} and~\ref{fig:qual_real}, we can conclude that the editing is preserving information from the source image while considering the edit prompt correctly.
In terms of preserving unrelated parts of images in the background, DALL-E 2 shows the best SSIM-$\overline{M}$ while \texttt{iEdit}-$M$ has the second best result.
InstructPix2Pix has more challenges in preserving the background.
Note that DALL-E 2 has an unfair advantage in our experiments, since it uses manually-provided masks at inference.
In terms of SSIM-$\overline{M}$, \texttt{iEdit}-$M$ outperforms all other methods and gets close to DALL-E 2 that uses ground-truth masks.

In Table~\ref{tab:quantitative_results_real}, we present a quantitative evaluation of our method and compared methods on editing the real images in our evaluation set. 
We observe a similar trend to the results obtained on generated images (Table~\ref{tab:quantitative_results_fake}). 
As before, even without using masks at inference, our method produces results with higher fidelity to the edit prompt in terms of CLIPScore. 
Conclusions similar to the case of generated images can be drawn for FID, SSIM-${M}$ and SSIM-$\overline{M}$.
To summarise, we conclude that our method is the best tradeoff between preserving the fidelity of the source image considering both foreground and background (SSIM and FFID) and being more aligned with the edit prompt (CLIPScore).

\subsection{Ablation Study}
Table~\ref{tab:quantitative_ablation} shows an ablation study that validates the effectiveness of each component of \texttt{iEdit} using the full evaluation set (generated and real images).
First, we compare LAION-caption-200K ($1^{st}$ row) and LAION-edit-200K ($2^{nd}$ row).
LAION-edit-200K is the dataset constructed as described in Sec.~\ref{sec:paireddataset} with BLIP-generated edit prompts, used as fine-tuning set in all previous experiments.
LAION-caption-200K is created in the same way but using the original captions from LAION-5B instead of the automatically generated ones. 
We use the best experimental setup for LAION-caption-200K, which employs $\mathcal{L}_{global} + \mathcal{L}_{pairs}$.
This is the best setup because the difference between these noisy source and target captions of LAION-5B are too high (even different languages sometimes) and often not grounded to images.
This prevents the use of masks in $\mathcal{L}_{mask}$ and of $\mathcal{L}_{perc}$ and $\mathcal{L}_{loc}$.
The $1^{st}$ and $2^{rd}$ rows of Table~\ref{tab:quantitative_ablation} show benefit of automatically-generated prompts 
with higher CLIPScore and SSIM-$\overline{M}$. 

Moreover, we ablate our main losses and components forming the masked image editing framework, by training only with the localised CLIP loss $\mathcal{L}_{loc}$ ($2^{nd}$ row), by adding $\mathcal{L}_{perc}$ ($3^{rd}$ row) and by switching $\mathcal{L}_{LDM}$ in favour of $\mathcal{L}_{mask}$ ($4^{th}$ row). 
The addition of $\mathcal{L}_{perc}$ marginally boosts the performance in all metrics,  while $\mathcal{L}_{mask}$ leads to a large improvement in CLIPScore and FID.
The final row of Table~\ref{tab:quantitative_ablation} that uses predicted masks at inference shows a large improvement in all metrics.
\section{Limitations}
Our method can be misguided during training by some automatically-generated pairs.
We observed that sometimes we pair very different visual content between images, making the editing task difficult.
InstructPix2Pix \cite{InstructPix2Pix} proposes a workaround to generate those images instead of retrieving them whereas employing pre-trained DMs to solve the task in a cyclic way.
To keep training accessible and feasible, we limited ourselves to low computational resources.
This can be lifted to provide better results for image editing, but with higher financial costs and $CO_2$ footprint.
Evaluation of image editing methods is one of the challenges due to missing ad-hoc metrics and non-existing common evaluation sets.
Since human evaluation is costly and subjective, we plan to investigate a more proper evaluation in the future.


\section{Conclusion}
We presented \texttt{iEdit}, a novel method for text-guided image editing based on LDMs. 
In addition, we proposed a method to automatically create a dataset train our model with weak supervision.
We introduced loss functions that use masks to enable localised preservation and synthesis of semantics and regions for editing.
Our model is fine-tuned on a small dataset (200K vs. 170M samples for LDMs), by using only 2 16GB GPUs.
\texttt{iEdit} achieves favourable qualitative and quantitative results against state-of-of-the-art methods on editing of generated and real images.

\vfill
\pagebreak

{\small
\bibliographystyle{ieee_fullname}
\bibliography{egbib}
}

\end{document}